\documentclass[a4paper, 9 pt, conference]{IEEEtran}
\IEEEoverridecommandlockouts    

\usepackage{gensymb}
\usepackage{enumerate}
\usepackage{graphicx}
\usepackage{subcaption}
\usepackage{color,soul}
\usepackage{capt-of}
\usepackage{flushend}
\usepackage{dblfloatfix}
\usepackage{xcolor,soul}
\usepackage{soul}
\usepackage{balance}

\AtBeginDocument{%
  \providecommand\BibTeX{{%
    \normalfont B\kern-0.5em{\scshape i\kern-0.25em b}\kern-0.8em\TeX}}}

\begin{document}

\title{Analysis of Power-Oriented Fault Injection Attacks on Spiking Neural Networks\vspace{-5mm}}

\author{%
Karthikeyan Nagarajan\textsuperscript{$\ast$}, Junde Li\textsuperscript{$\ast$}, Sina Sayyah Ensan\textsuperscript{$\ast$} \\Mohammad Nasim Imtiaz Khan\textsuperscript{$\dagger$}, Sachhidh Kannan\textsuperscript{$\ddagger$}, and Swaroop Ghosh\textsuperscript{$\ast$}\\
School of EECS, Pennsylvania State University, Univeristy Park, PA, USA\textsuperscript{$\ast$}\\  
Intel Corporation, Folsom, CA, USA\textsuperscript{$\dagger$}, 
Ampere Computing, Portland, OR, USA\textsuperscript{$\ddagger$}  \\ $\{$kxn287, jul1512, sayyah, szg212$\}$@psu.edu\textsuperscript{$\ast$} mohammad.nasim.imtiaz.khan@intel.com\textsuperscript{$\dagger$} sachhidh@amperecomputing.com\textsuperscript{$\ddagger$}%
\vspace{-4mm}
}

\maketitle

\begin{abstract}
Spiking Neural Networks (SNN) are quickly gaining traction as a viable alternative to Deep Neural Networks (DNN). In comparison to DNNs, SNNs are more computationally powerful and provide superior energy efficiency. SNNs, while exciting at first appearance, contain security-sensitive assets (e.g., neuron threshold voltage) and vulnerabilities (e.g., sensitivity of classification accuracy to neuron threshold voltage change) that adversaries can exploit. We investigate global fault injection attacks by employing external power supplies and laser-induced local power glitches to corrupt crucial training parameters such as spike amplitude and neuron's membrane threshold potential on SNNs developed using common analog neurons.
We also evaluate the impact of power-based attacks on individual SNN layers for 0\% (i.e., no attack) to 100\% (i.e., whole layer under attack). We investigate the impact of the attacks on digit classification tasks and find that in the worst-case scenario, classification accuracy is reduced by 85.65\%. We also propose defenses e.g., a robust current driver design that is immune to power-oriented attacks, improved circuit sizing of neuron components to reduce/recover the adversarial accuracy degradation at the cost of negligible area and 25\% power overhead. We also present a dummy neuron-based voltage fault injection detection system with $\sim$1\% power and area overhead.

\end{abstract}

\maketitle

\section{Introduction}

Artificial Neural Networks (ANNs or NNs), which are inspired by human brain functionality, are composed of layers of neurons interlinked by synapses and can be used to approximate any computable function. The use of neural networks in safety-critical domains, such as autonomous driving \cite{kaiser2016towards}, healthcare \cite{rahimiazghadi2020hardware}, Internet of Things \cite{whatmough2018dnn} and security \cite{cao2015spiking}, necessitates an examination of their security vulnerabilities and risks. In real-world applications, attacking a neural network can result in undesirable or dangerous inferences (e.g., reduced accuracy or confidence in road sign identification during autonomous driving). These attacks can be launched during the training, manufacturing, or final application stages.

Spiking Neural Networks (SNNs) \cite{maass1997networks} are the third generation of neural networks. SNNs are emerging as an alternative to Deep Neural Networks (DNNs) since they are biologically plausible, computationally powerful \cite{heiberg2018firing}, and energy-efficient \cite{merolla2014million}\cite{davies2018loihi}\cite{tavanaei2019deep}. However, very limited research exists on the the security of SNNs against adversarial attacks. Broadly, the attacks could be classified as: White Box attacks where an attacker has complete knowledge of the SNN architecture, and Black Box attacks where the attacker does not know the SNN architecture, network parameters or training data.

Multiple prior works \cite{goodfellow2014explaining}\cite{kurakin2016adversarial}\cite{madry2017towards} investigate adversarial attacks on DNN e.g., undetectable modifications to input data, causing a classifier to mispredict with an increased probability and propose countermeasures. The vulnerabilities/attacks of SNNs under a white-box scenario e.g., sensitivity to adversarial examples and a robust training mechanism for defense is proposed \cite{bagheri2018adversarial}.  
A white-box fault injection attack is proposed \cite{venceslai2020neuroattack} for SNNs by employing adversarial input noise. In \cite{marchisio2020spiking}, a black-box approach is presented to generate adversarial input instances to induce misprediction in SNNs. However, the impact of voltage/power-based fault injection attacks are not studied in these white- and black-box SNN attack scenarios.

In prior works, Voltage fault injection (VFI) techniques have been proven to be powerful side channel attacks for disrupting a system's execution flow. In {\cite{barenghi2012fault}}, a fault injection approach is described that underpowers cryptographic devices and introduces bit errors. In {\cite{bozzato2019shaping}}, novel VFI techniques are proposed to inject glitches in popular microcontrollers from manufacturers such as, STMicroelectronics and Texas Instruments. In {\cite{zussa2013power}}, timing constraint violations were introduced in FPGAs by using a negative power supply glitch attack. Laser-based injection has also been proposed for inducing local voltage and clock glitching attacks. However, such studies are not performed for SNN.

\begin{figure} [t] 
 \vspace{-3mm}
 \begin{center}
    \includegraphics[width=.45\textwidth]{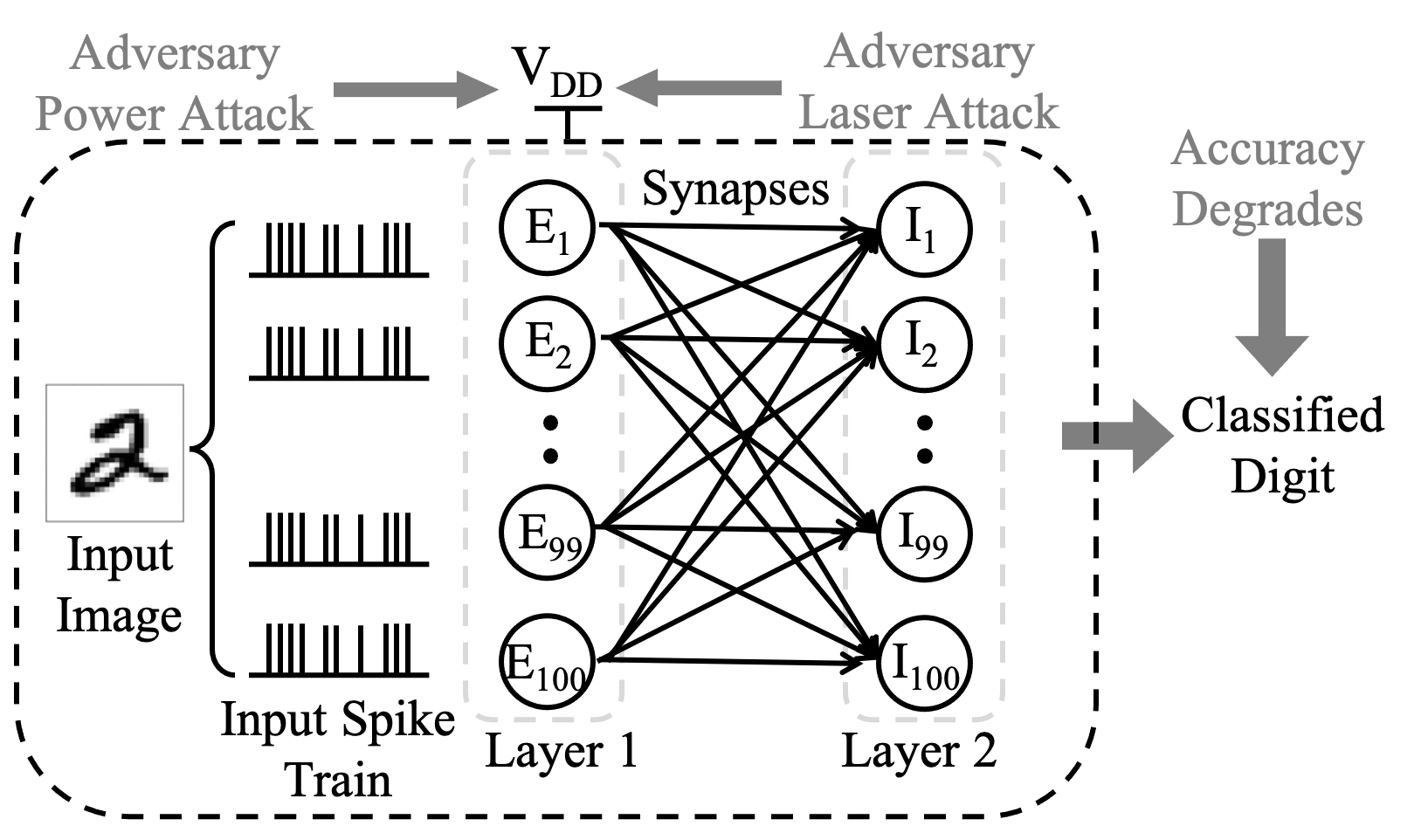}
 \end{center}
\vspace{-4mm}
 \caption{Threat model for power-based attacks on SNN.} 
 \label{threat_model}
\vspace{-4mm}
 \end{figure}




\textbf{Proposed Threat Model:} There is limited amount of research on SNN attacks (except adversarial input-based attacks). Similar to classical systems, the adversary can manipulate the supply voltage or inject voltage glitches in the SNN systems. This is likely for, (i) an external adversary who has physical possession of the device or the power port, (ii) an insider adversary with access to power port or laser gun to inject the fault. In this paper, we study a total of five attacks under both black box and white box scenarios. 

\emph{Black Box Attack: }In this scenario (Attack-5 in Section {\ref{BB}}), the adversary affects the power supply of the entire system to (i) corrupt spiking amplitude of SNN neuron input and, (ii) disrupt SNN neuron's membrane functionality. To launch this attack, the adversary does not need to know the SNN architecture but needs the control of the external power supply ($V_{DD}$). Fig. {\ref{threat_model}} shows a high-level schematic of the proposed threat model against an SNN, where an input image to be classified is converted to spike trains and fed to the neuron layers. The objective is to degrade accuracy of the classified digit. Note that the neuron layers, neurons, and  interconnections shown in Fig. {\ref{threat_model}} just illustrate the proposed threat model. The SNN architecture actually implemented in this paper is explained in Section {\ref{exp_setup}}. 

\begin{figure*} [t] 
 \vspace{-1mm}
 \begin{center}
    \includegraphics[width=.92\textwidth]{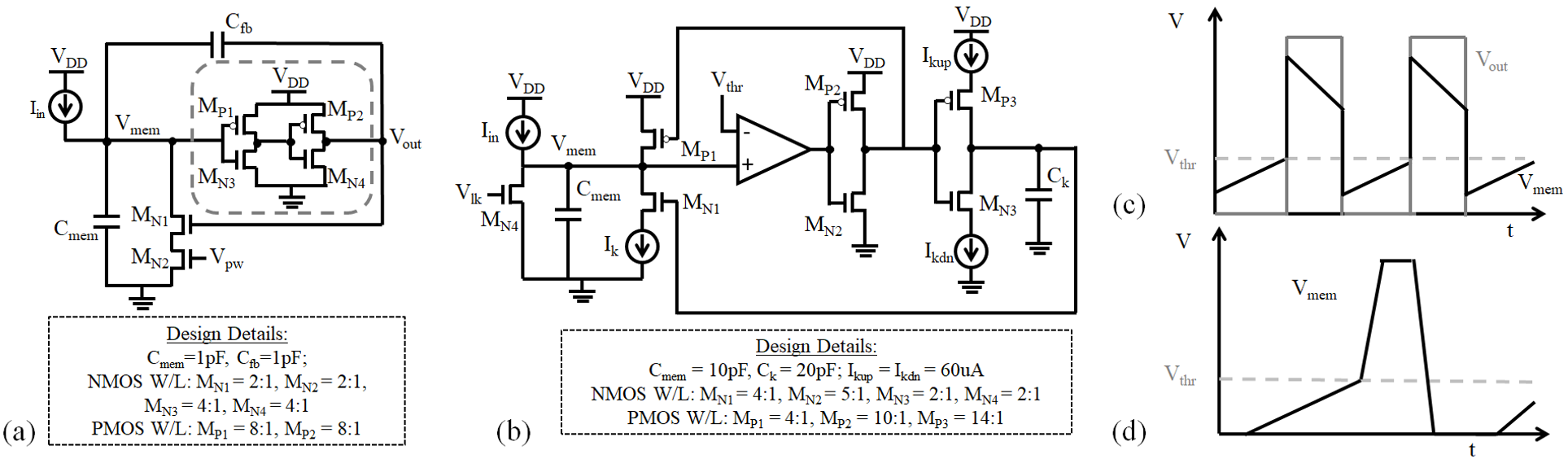}
 \end{center}
 \vspace{-3mm}
 \caption{(a) Axon Hillock circuit; (b) Voltage amplifier I\&F circuit; (c) Expected membrane voltage and output voltage of Axon Hillock neuron; (d) Expected membrane voltage voltage of I\&F neuron.} 
 \label{ind_neurons}
\vspace{-3mm}
 \end{figure*}
 
\emph{White Box Attacks:} In this scenario, we consider following cases (details in Sections {\ref{WB-1}} and {\ref{WB-2}}) where the adversary is able to individually attack SNN layers and peripherals through localized laser based power fault injection, (i) Attack 1 where only peripherals e.g., input current drivers are attacked, (ii) Attack 2 and 3 where individual SNN layers attacked partially to fully i.e., 0\%-100\%, and (iii) Attack 4 where all SNN layers affected (no peripherals). 

\textbf{Contributions}: In summary, we, (a) present detailed analysis of two neuron models namely, Axon Hillock neuron and voltage I\&F amplifier neuron under global, local and fine-grain supply voltage variation; (b) propose five power-based attack models against SNN designs under black box and white box settings; (c) analyze impact of proposed attacks for digit classification tasks; and, (d) propose defenses and a novel detection technique.



In the remaining of the paper, Section II presents background on SNNs and neuron design, Section III proposes the attack models, Sections IV and V present the analysis of the attack and countermeasures, respectively and finally, Section VI draws the conclusion.

\section{Background}
In this section, we present the overview of SNN and neuron designs \cite{indiveri2011neuromorphic} that have been used in this paper. 
\subsection{Overview of Spiking Neural Network}

SNNs are composed of layers of spiking neurons that are interconnected together by synaptic weights (Fig. \ref{threat_model}). The neurons between adjacent layers exchange information in the form of spike trains. The timing of the spikes and the strength of the synaptic weights between neurons are critical parameters in SNN operation. Each neuron includes a membrane, whose potential increases when the neuron receives an input spike. The neuron \emph{fires} an output spike when this membrane potential crosses a pre-determined threshold. Various neuron models such as, I\&F, Hodgkin-Huxley, and spike response exist with different membrane and spike-generation operations. In this work, we have implemented two flavors of I\&F neuron to showcase the power-based attacks.  


\subsection{Neuron Design and Implementation} \label{SNN}

In this work, we implement, simulate, and analyze all neuron models on HSPICE using PTM 65nm technology.

\subsubsection{Axon Hillock Spiking Neuron Design}
 The Axon Hillock circuit \cite{mead2012analog} (Fig. \ref{ind_neurons}a) consists of an amplifier block implemented using two inverters in series (shown in dotted gray box). The input current ($I_{in}$) is integrated at the neuron membrane capacitance ($C_{mem}$), and the analog membrane voltage ($V_{mem}$) rises linearly until it crosses the amplifier's threshold. Once it reaches this point, the output  ($V_{out}$) switches from `0' to $V_{DD}$. This $V_{out}$ is fed back into a reset transistor ($M_{N1}$) and activates a positive feedback through the capacitor divider ($C_{fb}$). Another transistor ($M_{N2}$), controlled by $V_{pw}$, determines the reset current. If reset current $>$ $I_{in}$, $C_{mem}$ is discharged until it falls to the amplifier's threshold. This causes $V_{out}$ to switch from $V_{DD}$ to `0'. The output remains `0' until the entire cycle repeats. Fig. \ref{ind_neurons}c depicts the expected results of $V_{mem}$ and $V_{out}$.

\begin{figure} [t] 
 \vspace{-1mm}
 \begin{center}
    \includegraphics[width=.27\textwidth]{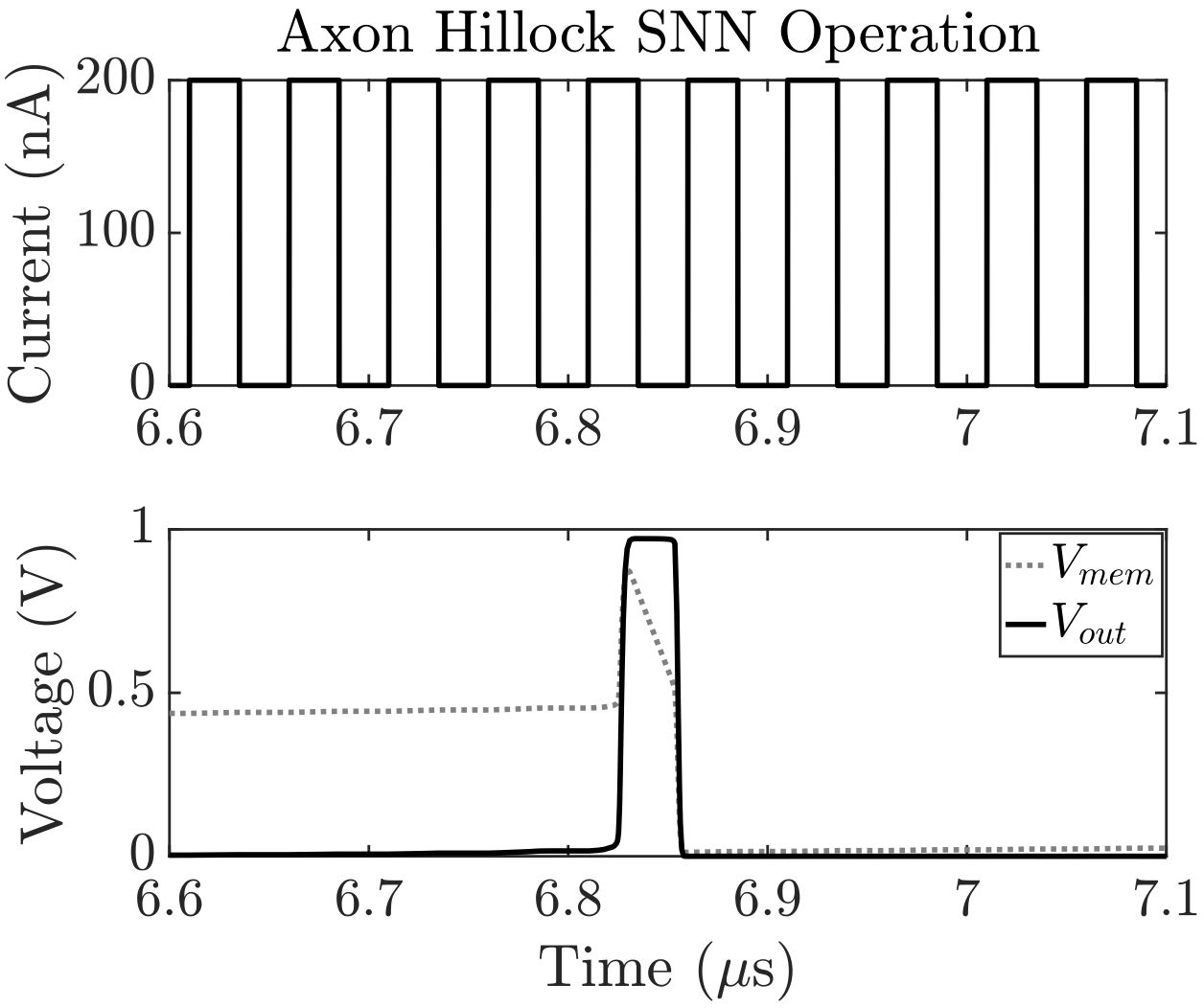}
 \end{center}
 \vspace{-3mm}
 \caption{Simulation result of Axon Hillock spike generation showing input current ($I_{in}$) (top plot), the membrane voltage ($V_{mem}$) and the output voltage ($V_{out}$) (bottom plot).} 
 \label{axon_hillock_sim}
\vspace{-3mm}
 \end{figure}

In this paper, the value of membrane capacitance ($C_{mem}$) and the feedback capacitance ($C_{fb}$) of 1pF are used. For experimental purposes, the input current spikes with an amplitude of 200nA, a spike width of 25ns, and a spike rate of 40MHz are generated through the current source ($I_{in}$). The $V_{DD}$ of the design is set to 1V. Fig. \ref{axon_hillock_sim} shows the simulation results of the  input current spikes ($I_{in}$) and the corresponding membrane and the output voltage ($V_{out}$).


    \begin{figure} [t] 
 \vspace{-1mm}
 \begin{center}
    \includegraphics[width=.265\textwidth]{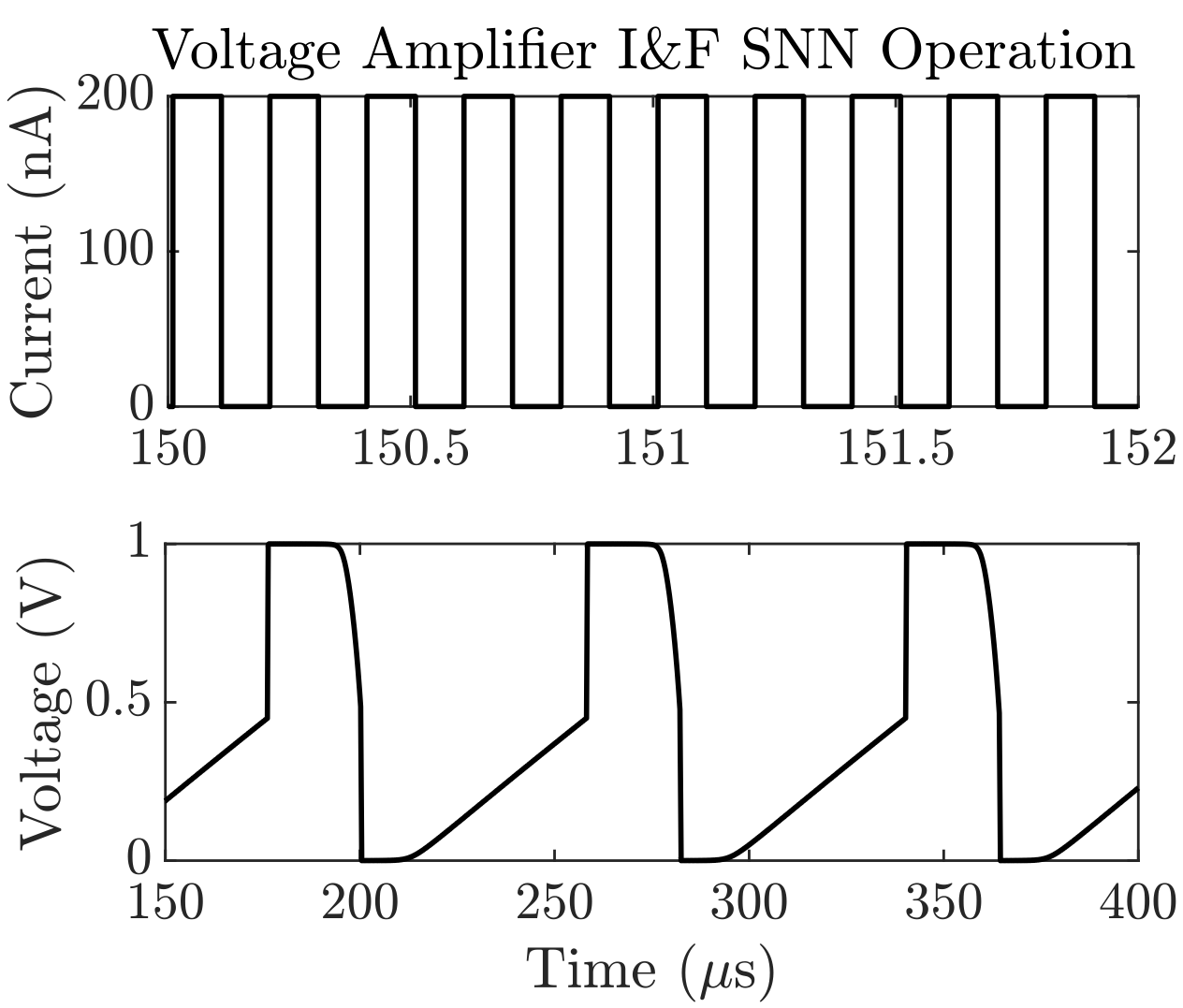}
 \end{center}
 \vspace{-3mm}
 \caption{Simulation result of voltage amplifier I\&F neuron spike generation showing input current ($I_{in}$) (top plot and zoomed-in), and the membrane voltage ($V_{mem}$) (bottom plot).} 
 \label{IF_sim}
\vspace{-5mm}
 \end{figure}

  \begin{figure*} [t] 
        \centering 
        (a)
        \begin{subfigure}[b]{0.27\textwidth}
                \centering
                \includegraphics[width=0.99\linewidth]{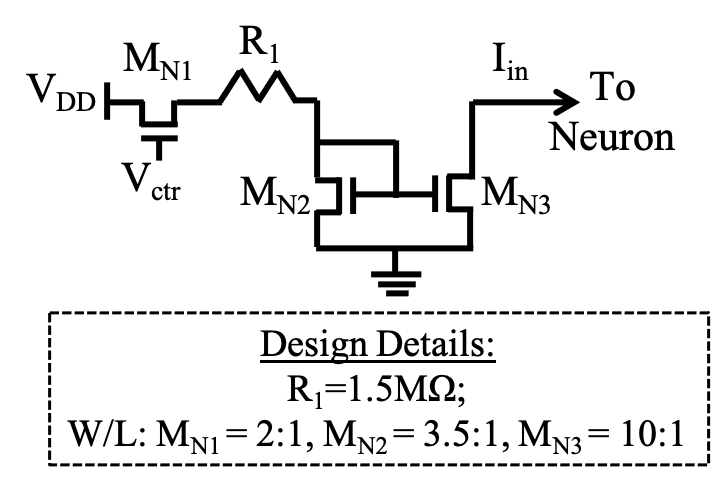}
                \vspace{-9mm}
                \caption{}
                \label{cur_driver}
        \end{subfigure}%
         \hspace{0mm} (b)
        \hspace{0mm}
        \begin{subfigure}[b]{0.27\textwidth}
                \centering
                \includegraphics[width=0.99\linewidth]{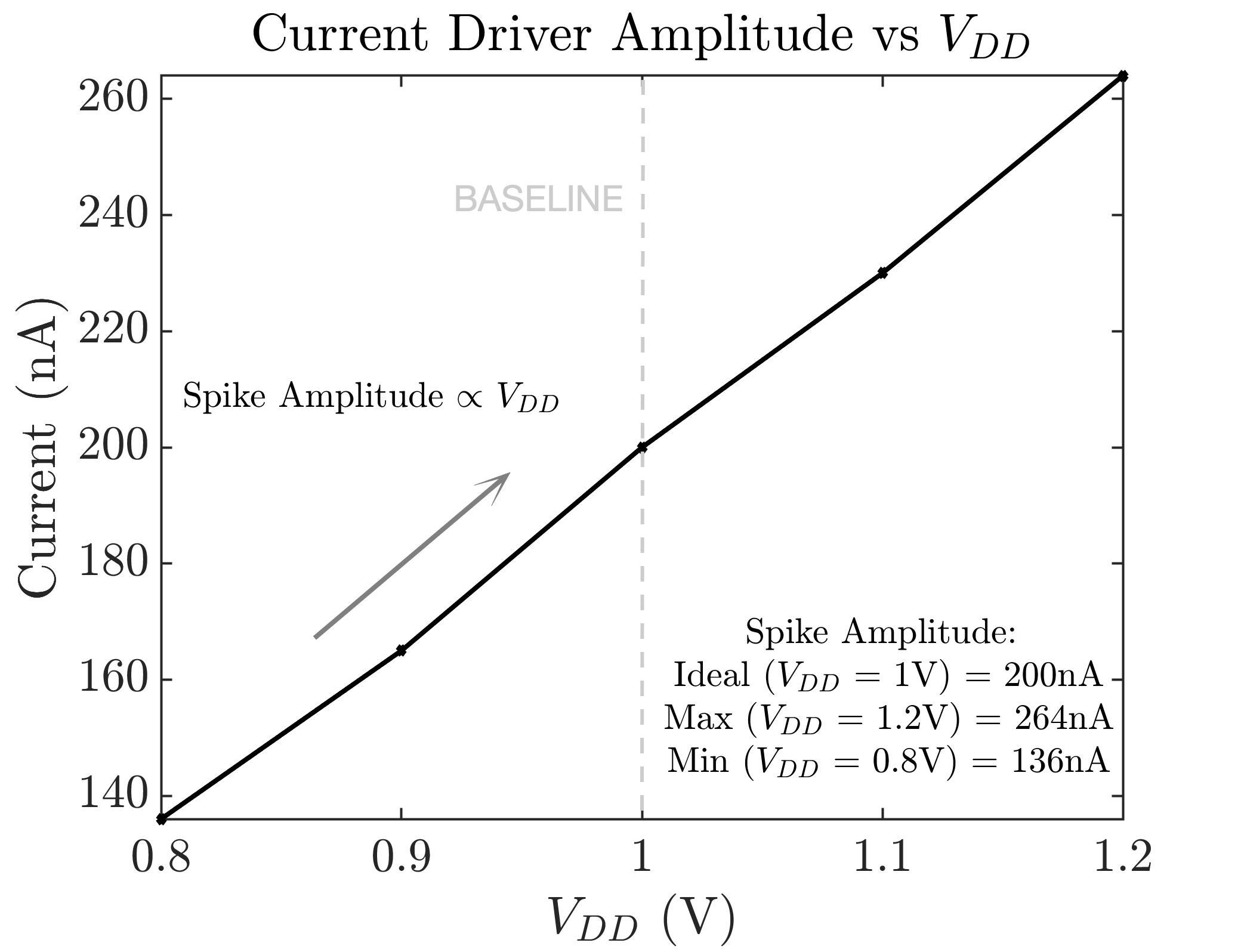}
                \vspace{-11mm}
                \caption{}
                \label{cur_driver_amp}
        \end{subfigure}%
         \hspace{0mm} (c)
        \hspace{-2mm} 
        \begin{subfigure}[b]{0.27\textwidth}
                \centering
              \includegraphics[width=0.99\linewidth]{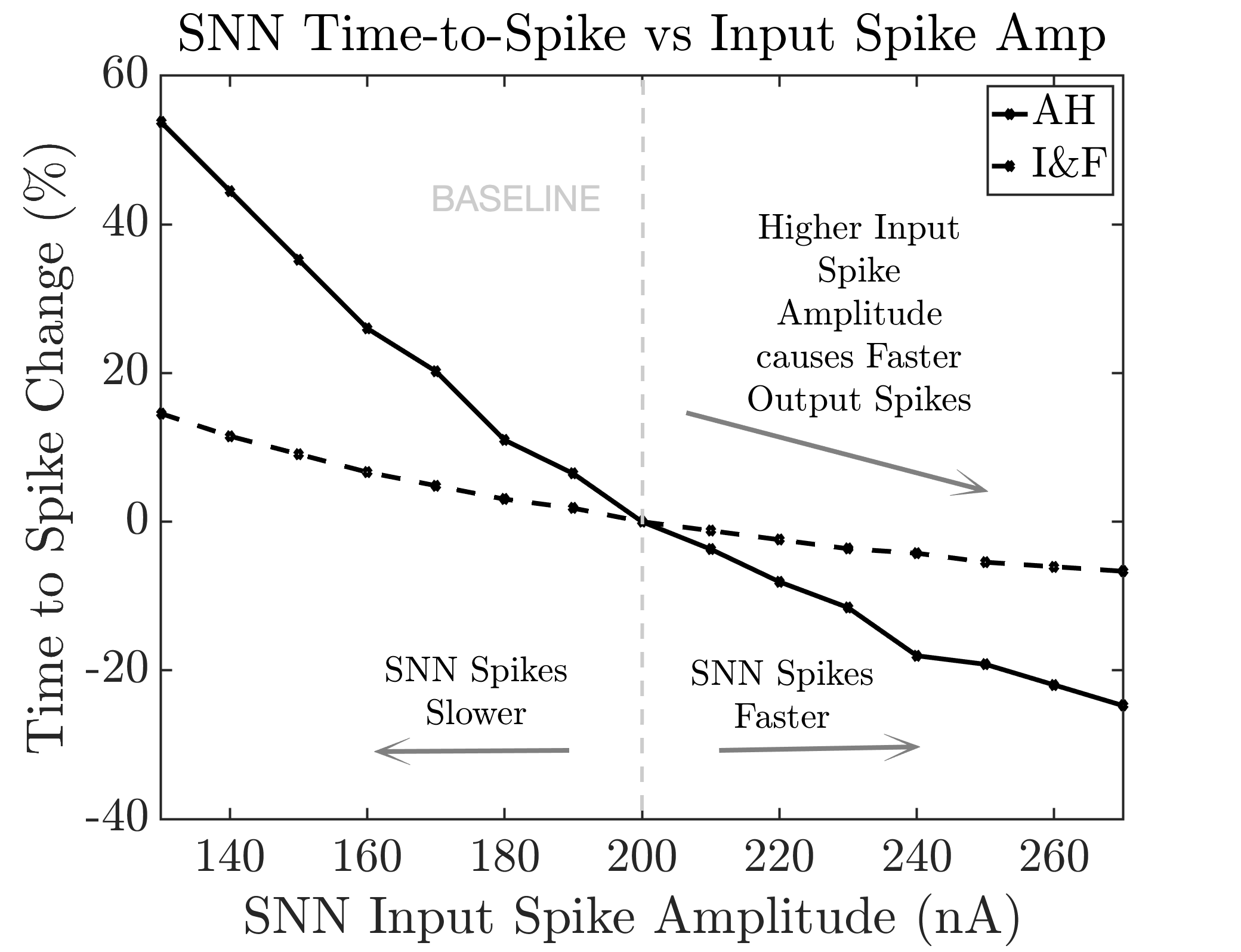}
                 \vspace{-11mm}
                \caption{}
                \label{snn_both_amp}
        \end{subfigure}%
        \vspace{-0mm}
        \caption{(a) Current driver circuit of the SNN neurons with design details; (b) Change in current driver output spike ($I_{in}$) amplitude with change in $V_{DD}$; and (C) Effect of input spike amplitude on SNN output time-to-spike for Axon Hillock neuron and voltage amplifier I\&F neuron.}
        \label{snac1}
        \vspace{-2mm}
\end{figure*}

 \subsubsection{Voltage Amplifier I\&F Neuron Design}
The voltage amplifier I\&F circuit \cite{van2001building} (Fig. \ref{ind_neurons}b) employs a 5-transistor amplifier that offers better control over the threshold voltage of the neuron. This design allows the designer to determine an explicit threshold and an explicit refractory period.  The threshold voltage ($V_{thr}$) of the amplifier employed is set to 0.5V and the $V_{DD}$ is set to 1V. The neuron membrane is modeled using a 10pF capacitance ($C_{mem}$) and the membrane leakage is controlled by transistor $M_{N4}$ with a gate ($V_{lk}$) voltage of 0.2V. The excitatory input current spikes ($I_{in}$) integrates charge over $C_{mem}$ and the node voltage at $V_{mem}$ rises linearly. Once $V_{mem}$ crosses $V_{thr}$, the comparator output switches from `0' to $V_{DD}$. This output is fed into 2 inverters in series, where the output of the first inverter is used to pull up $V_{mem}$ to $V_{DD}$ and the output of the second inverter is used to charge a second capacitor ($C_{k}$) of 20pF. The node voltage of $C_{k}$ is fed back to a reset transistor $M_{N1}$. When this node voltage is high enough, $M_{N1}$ is activated and $V_{mem}$ is pulled down to `0' and remains LOW until $C_{k}$ discharges below the activation voltage of $M_{N1}$. For experimental purposes, the input current spikes with an amplitude of 200nA, a spike width of 25ns, and a time interval of 25ns between consecutive spikes are generated through the current source ($I_{in}$). Fig. \ref{ind_neurons}d depicts the expected results of $V_{mem}$. Fig. \ref{IF_sim} shows the simulation results of input current spikes ($I_{in}$) and corresponding membrane voltage ($V_{out}$).


  \begin{figure*} [t] 
        \centering 
        (a)
        \begin{subfigure}[b]{0.27\textwidth}
                \centering
                \includegraphics[width=0.99\linewidth]{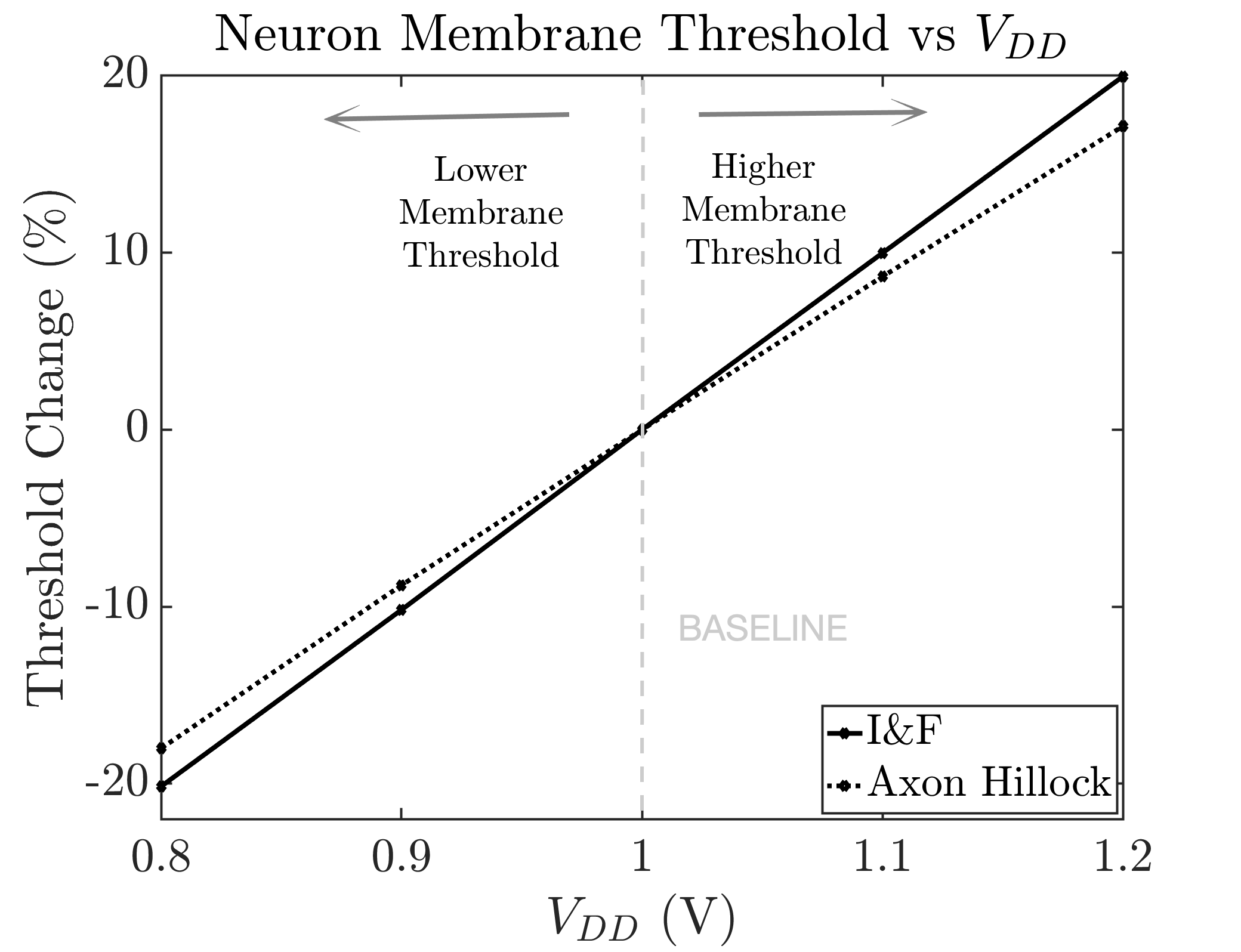}
                 \vspace{-11mm}
                \caption{}
                \label{snn_thresh_vdd}
        \end{subfigure}%
        \hspace{0mm} (b)
        \begin{subfigure}[b]{0.27\textwidth}
                \centering
                \includegraphics[width=0.99\linewidth]{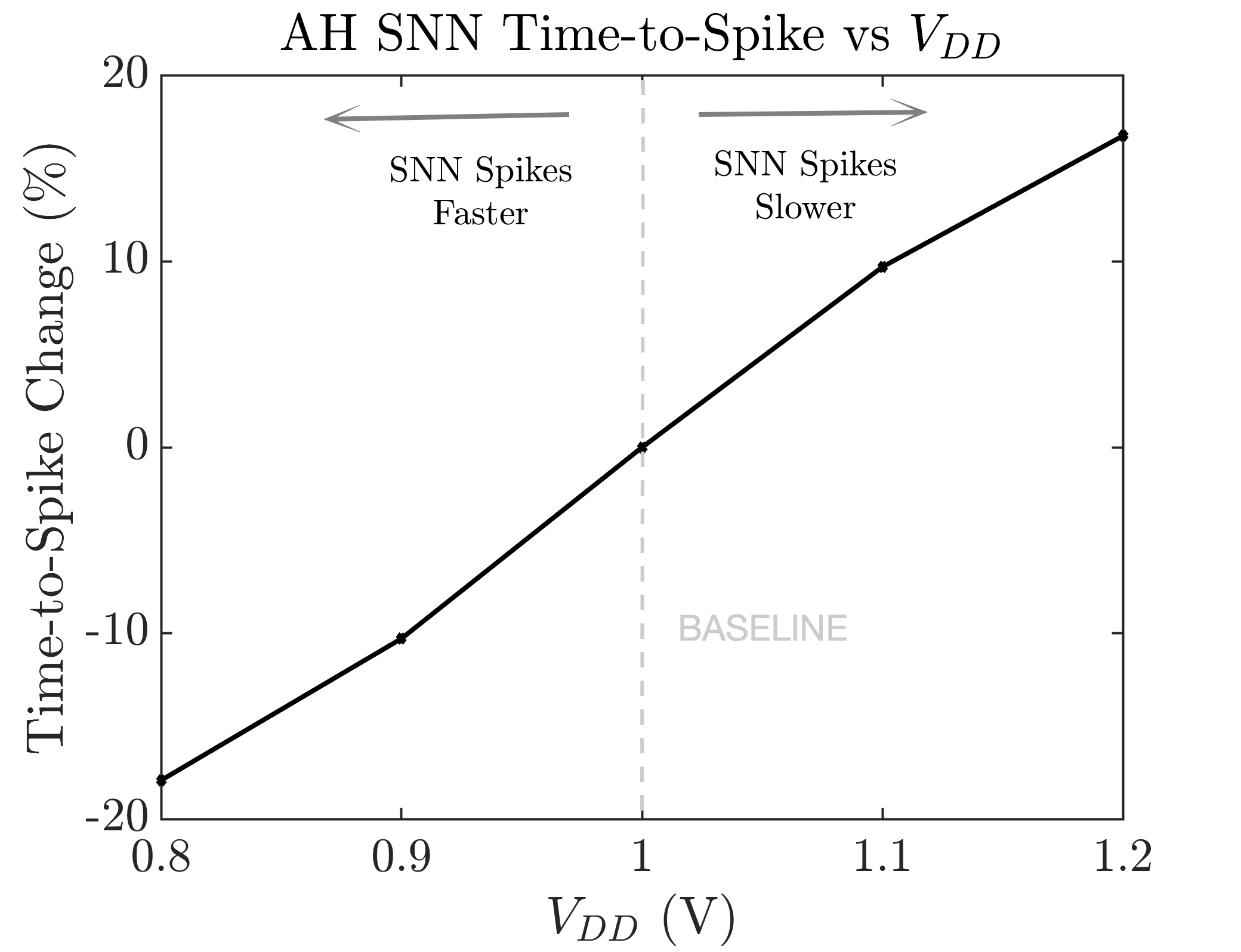}
                 \vspace{-11mm}
                \caption{}
                \label{spike_axon_vdd}
        \end{subfigure}%
        \hspace{0mm} (c)
        \begin{subfigure}[b]{0.27\textwidth}
                \centering
              \includegraphics[width=0.99\linewidth]{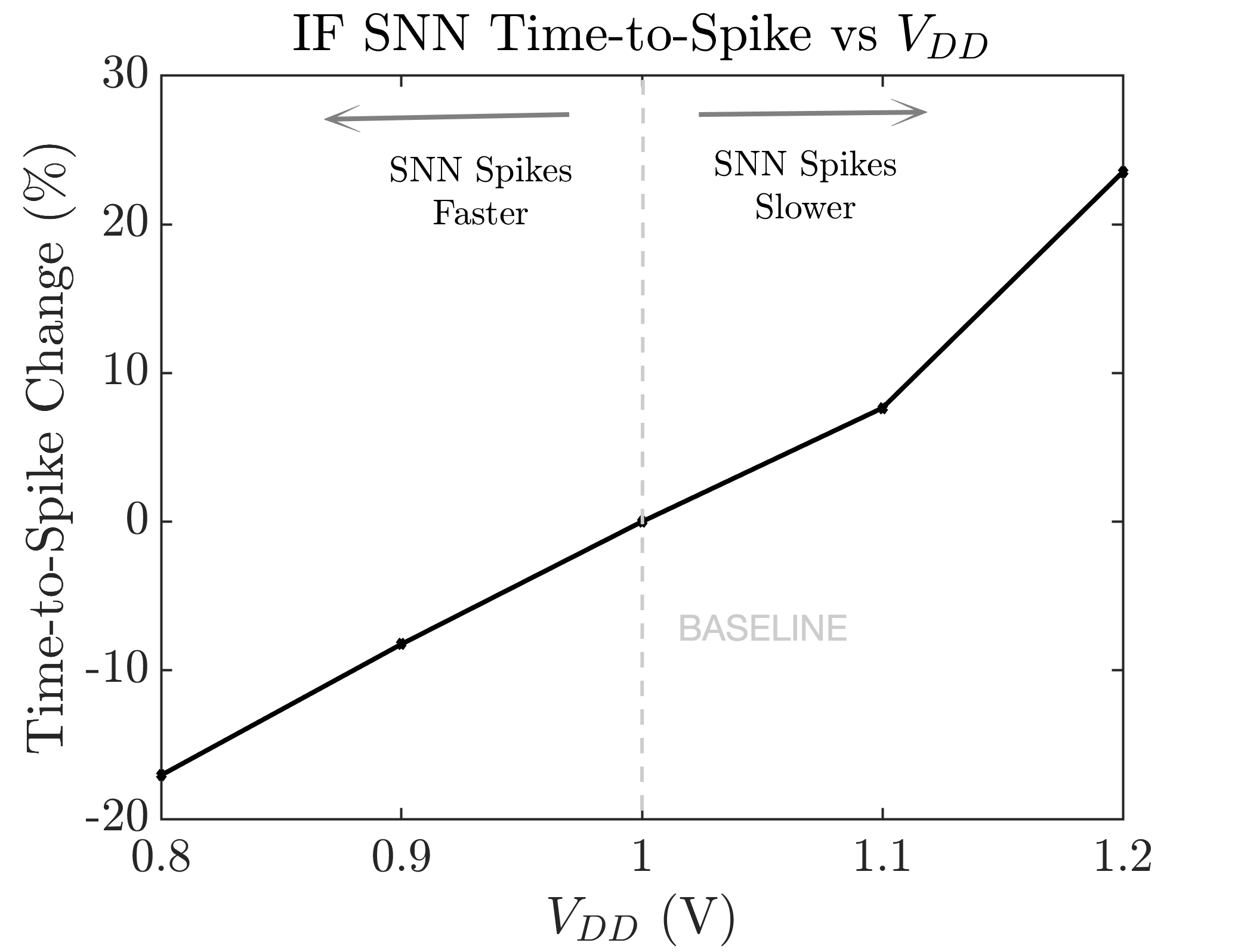}
                  \vspace{-11mm}
                \caption{}
                \label{spike_IF_vdd}
        \end{subfigure}%
        \vspace{-0mm}
        \caption{(a) Change in SNN membrane threshold with change in $V_{DD}$; Effect of $V_{DD}$ change on SNN output time-to-spike for (b) Axon Hillock neuron; (c) Voltage amplifier I\&F neuron.}
        \label{snac2}
        \vspace{-2mm}
\end{figure*}

 \subsection{SNN Current Driver Design} \label{cur_drive}
 A current driver provides the input current spikes to the neuron e.g., image input converted to current spike train. We have designed a current source based on a current mirror (Fig. \ref{cur_driver}) where $V_{GS}$ of $M_{N2}$ and  $M_{N3}$ are equal causing both transistors to pass the same current. The sizes of the $M_{N2}$ and $M_{N3}$ transistors and the resistor ($R_1$) are chosen to provide a current of amplitude 200 nA. Since the input current of the neuron is modeled as spikes, we have added the $M_{N1}$ transistor to act as a switch that is controlled by incoming voltage spikes ($V_{ctr}$) from other neurons.
 

\section{Neuron Attack Models} \label{attack_models}
In this section, we describe the power-based attacks and examine the impact of $V_{DD}$ manipulation on crucial circuit components and parameters of the previously explained SNNs.

\subsection{Attack Assumptions} \label{attack_assumptions}
We have investigated the power attacks under the following cases:
\subsubsection{Case 1: Separate Power Domains} 
The current drivers and neurons (of the entire SNN) are assumed to be operated on separate $V_{DD}$ domains. This is possible if the neurons, synapses and peripherals have distinct supply voltages e.g., if the neuron and peripherals are CMOS and the synapses are based on memristers. This case enables us to study the effect of $V_{DD}$ modulation on individual components.

\subsubsection{Case 2: Single Power Domain} The entire SNN system, including current drivers and neurons share the same $V_{DD}$. This is a likely scenario if the whole circuit is CMOS based.

\subsubsection{Case 3: Local Power Glitching} The adversary has fine grain control of the $V_{DD}$ inside a voltage domain for both separate and single power domain cases. For example, adversary can use a focused laser beam to cause localized voltage glitching.


\subsection{SNN Input Spike Corruption} \label{snac1_des}
The input current spikes of each neuron are fed using a current driver as described in Section \ref{cur_drive}. The driver is designed with $V_{DD}$ = 1V and outputs SNN input current spikes of 200nA amplitude and 25ns spike width. 
An adversary can attack a normal driver operation by modulating the $V_{DD}$. 

Fig. \ref{cur_driver_amp} shows the effect of modulating the $V_{DD}$ from 0.8V to 1.2V (corresponding to a  -/+ 20\% change). The corresponding output spike amplitude ranges from 136nA for 0.8$V_{DD}$ (-32\% change) to 264nA for 1.2$V_{DD}$ (+32\% change). We subjected our neuron designs under these input spike amplitude modulations while keeping the input spiking rate constant at 40MHz. Fig. \ref{snn_both_amp} shows the effect on output spike rate for the Axon Hillock neuron where the time-to-spike ($V_{out}$) becomes faster by 24.7\% under $V_{DD}$=1.2V and $I_{in}=264nA$ and becomes slower by 53.7\% under $V_{DD}$=0.8V and $I_{in}=136nA$. Similarly, Fig. \ref{snac1}c also shows the effect on output spike rate for the voltage amplifier I\&F neuron where the time-to-spike ($V_{out}$) becomes faster by 6.7\% under $V_{DD}$=1.2V and $I_{in}=264nA$ and becomes slower by 14.5\% under $V_{DD}$=0.8V and $I_{in}=136nA$.

   \begin{figure*} [t] 
        \centering 
        (a)
        \begin{subfigure}[b]{0.57\textwidth}
                \centering
                \includegraphics[width=0.99\linewidth]{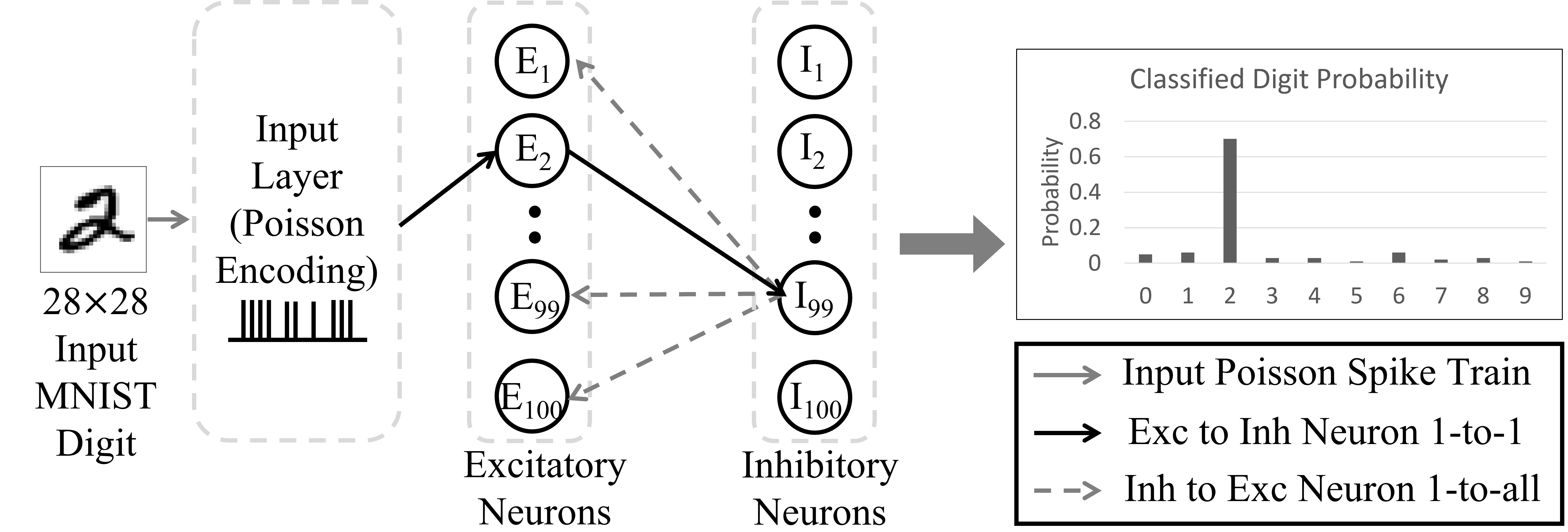}
                 \vspace{-5mm}
                \caption{}
                \label{bindsnet_imp}
        \end{subfigure}%
        \hspace{0mm}
        \hspace{0mm} (b)
        \begin{subfigure}[b]{0.27\textwidth}
                \centering
                \includegraphics[width=0.99\linewidth]{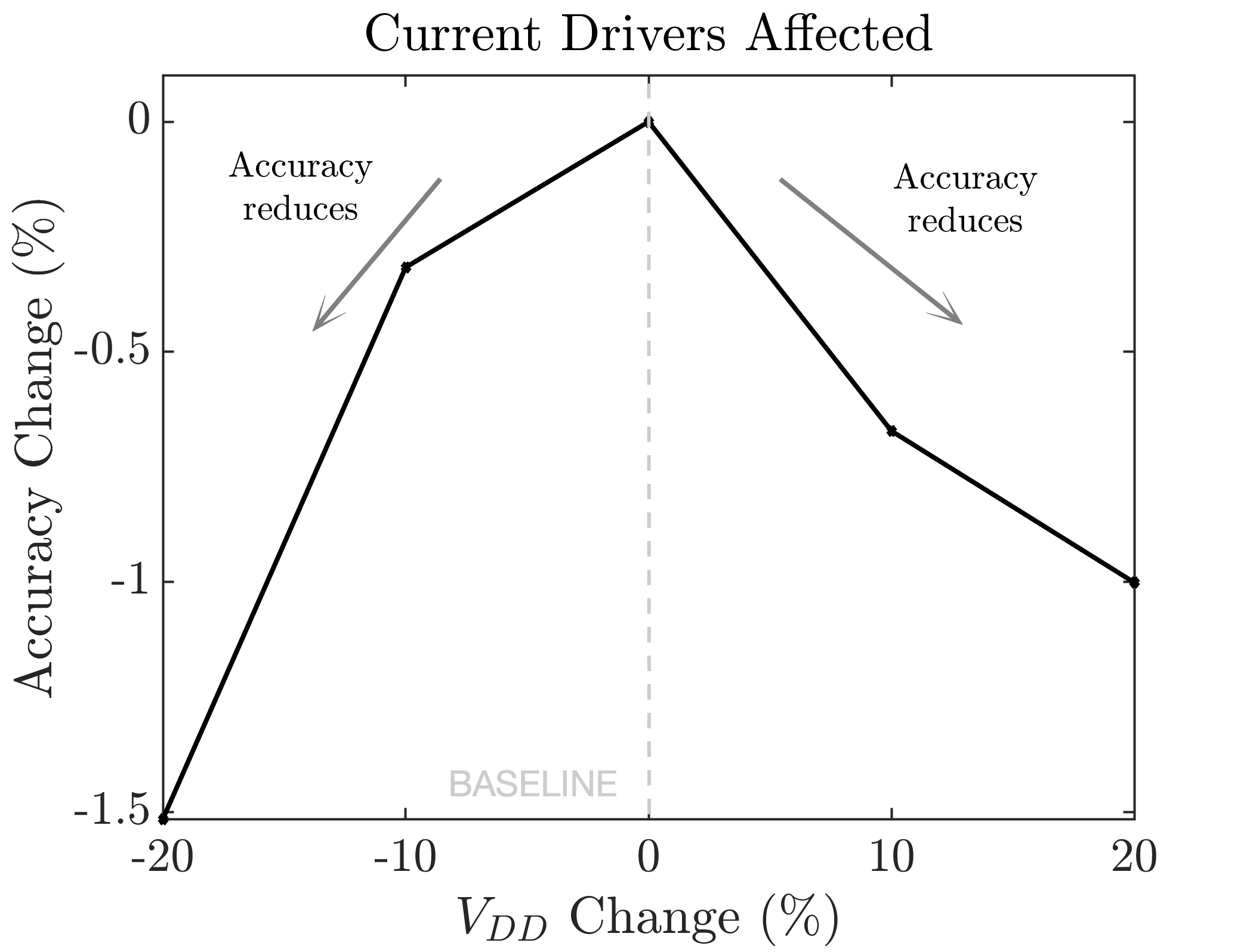}
                 \vspace{-11mm}
                \caption{}
                \label{theta_change}
        \end{subfigure}%
        \vspace{-0mm}
        \caption{(a) Implemented 3 layer SNN \cite{diehl2015fast}; (b) Effect of current driver corruption (Attack 1) on MNIST classification accuracy.}
        \label{binds_theta}
        \vspace{-2mm}
\end{figure*}

\subsection{SNN Threshold Manipulation} \label{snac2_des}
The adversary can also corrupt normal SNN operation using the externally supplied $V_{DD}$ which can modulate the SNN's membrane threshold voltage. In the ideal condition, the $V_{DD}$ is 1V and the threshold voltage of both the Axon Hillock neuron and the I\&F neuron are designed to be 0.5V. 
Fig. \ref{snn_thresh_vdd} shows that the membrane threshold voltage changes with $V_{DD}$. In case of the Axon Hillock neuron the change in threshold ranges from -17.91\% for $V_{DD}=0.8V$ to +16.76\% for $V_{DD}=1.2V$. When $V_{DD}$ is modified, the switching threshold of the inverters in the Axon Hillock neuron is also proportionally affected. A lower (higher) $V_{DD}$ lowers (increases) the switching threshold of the inverters and leads to a faster (slower) output spike. Similarly, the change in threshold ranges from -18.01\% to +17.14\% when $V_{DD}$ is swept from 0.8V to 1.2V for the voltage amplifier I\&F neuron. Note that the change in threshold for the I\&F neuron is due to $V_{thr}$ signal (Fig. \ref{snn_thresh_vdd}) which is derived using a simple resistor-based voltage division of $V_{DD}$. Therefore, $V_{thr}$ scales linearly with $V_{DD}$.

The change in membrane threshold modulates the output spike rate of the affected SNN neurons. Fig. \ref{spike_axon_vdd} and Fig. \ref{spike_IF_vdd} show the change in time-to-spike under $V_{DD}$ manipulation while the input spikes ($I_{in}$) to the neuron are held at a constant amplitude of 200nA and a rate of 40MHz. The time-to-spike for Axon Hillock ranges from 17.91\% faster to 16.76\% slower. Similarly, the time-to-spike for I\&F neuron ranges from 17.05\% faster to 23.53\% slower.

 \section{Analysis of Power Attacks on SNN } \label{SNN_model}

This section 
describes the effect of power-oriented attacks on the image classification accuracy under the attack assumptions from Section {\ref{attack_assumptions}}.

\subsection{Experimental Setup} \label{exp_setup}
We have implemented the Diehl\&Cook SNN \cite{diehl2015fast} using the BindsNET \cite{hazan2018bindsnet} network library with PyTorch Tensor to test the effect of power-based attacks. The SNN is implemented with 3 neuron layers (Fig. \ref{bindsnet_imp}), namely Input layer, Excitatory Layer (EL), and Inhibitory Layer (IL). We employ this SNN for digit classification of the MNIST dataset which consists of digit images of pixel dimension 28$\times$28. Each input image is converted to Poisson-spike trains and fed to the Excitatory neurons in an all-to-all connection, where each input spike is fed to each Excitatory neurons. The Excitatory neurons are 1-to-1 connected with the Inhibitory neurons (Fig. \ref{bindsnet_imp}). 
Each neuron in the Inhibitory layer is in turn connected to all the neurons in the Excitatory layer, except the one it received a connection from. The architecture performs supervised learning. For our experiments, the EL and IL have 100 neurons each and all experiments are conducted on 1000 Poisson-encoded training images with fixed learning rates of 0.0004 and 0.0002 for pre-synaptic and post-synaptic events, respectively. The batch size is set to 32 and training samples are iterated only once as configured in \cite{hazan2018bindsnet}. Additional details on the neuron layers, learning method, and SNN parameters can be found in {\cite{hazan2018bindsnet}}. The baseline classification accuracy for attack-free SNN is 75.92\% with 1000 training images. 


\subsection{Input Spike Corruption } \label{WB-1}

In Section \ref{snac1_des}, it is shown that the adversary can manipulate the input spike amplitudes for the SNN neurons. This in turn changes the membrane voltage by a different rate for the same number of input spikes. This manipulation of the rate of change of membrane voltage changes the time-to-spike for the neuron (as shown in Fig. \ref{snn_both_amp}).

\textbf {Attack 1: }In order to translate this effect to our BindsNET SNN implementation, we have modified the rate of change of the neuron's membrane voltage using variable \emph{theta} which specifies the voltage change in the neuron membrane for each input spike. 
Fig. \ref{theta_change} shows the corresponding change in MNIST digit classification accuracy. Under the worst case \emph{theta} change of -20\%, classification accuracy decreases by 1.5\%. It is seen that the classification accuracy is not adversely affected by increasing/decreasing \emph{theta} since the accuracy remains within +/-2\% of the baseline accuracy. Note that this is a \emph{white box} attack since the adversary requires the location of the current drivers within the SNN (possible by invasive reverse engineering of a chip) to induce the localized fault. 

 \begin{figure*} [t] 
        \centering 
        (a)
        \begin{subfigure}[b]{0.27\textwidth}
                \centering
                \includegraphics[width=0.99\linewidth]{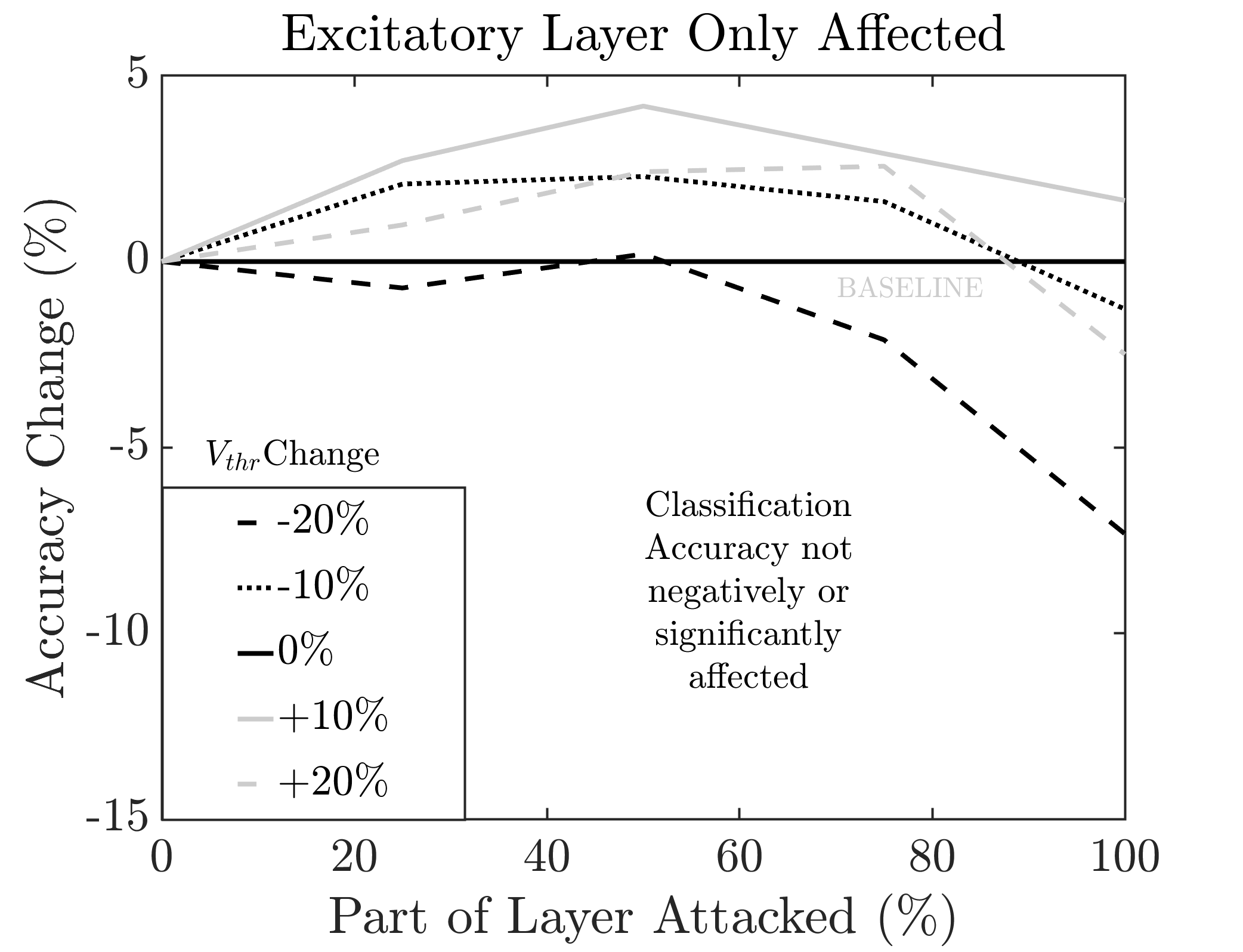}
                 \vspace{-11mm}
                \caption{}
                \label{EL_only}
        \end{subfigure}%
        \hspace{0mm} (b)
        \begin{subfigure}[b]{0.27\textwidth}
                \centering
                \includegraphics[width=0.99\linewidth]{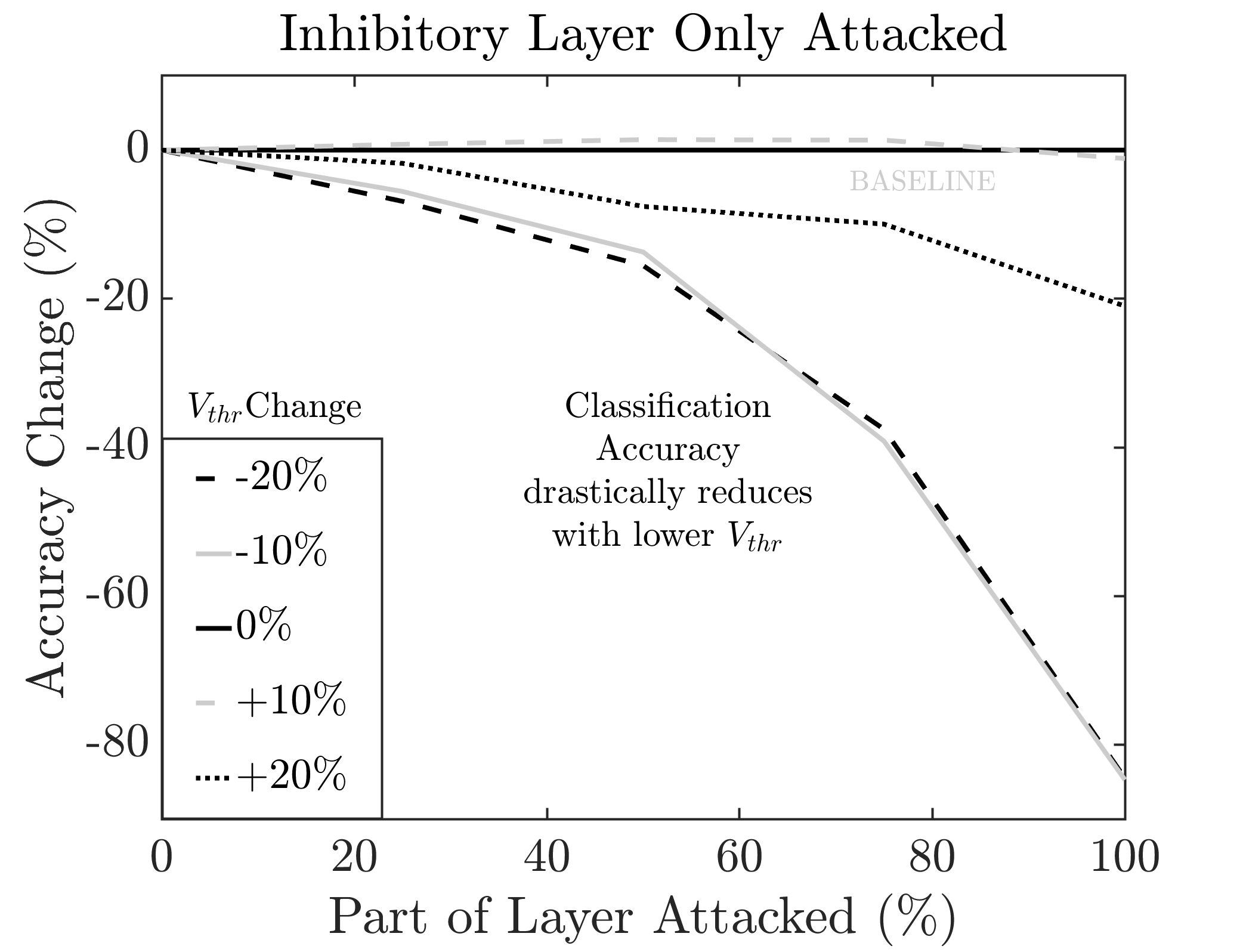}
                 \vspace{-11mm}
                \caption{}
                \label{IH_only}
        \end{subfigure}%
        \hspace{0mm} (c)
        \begin{subfigure}[b]{0.27\textwidth}
                \centering
               \includegraphics[width=0.99\linewidth]{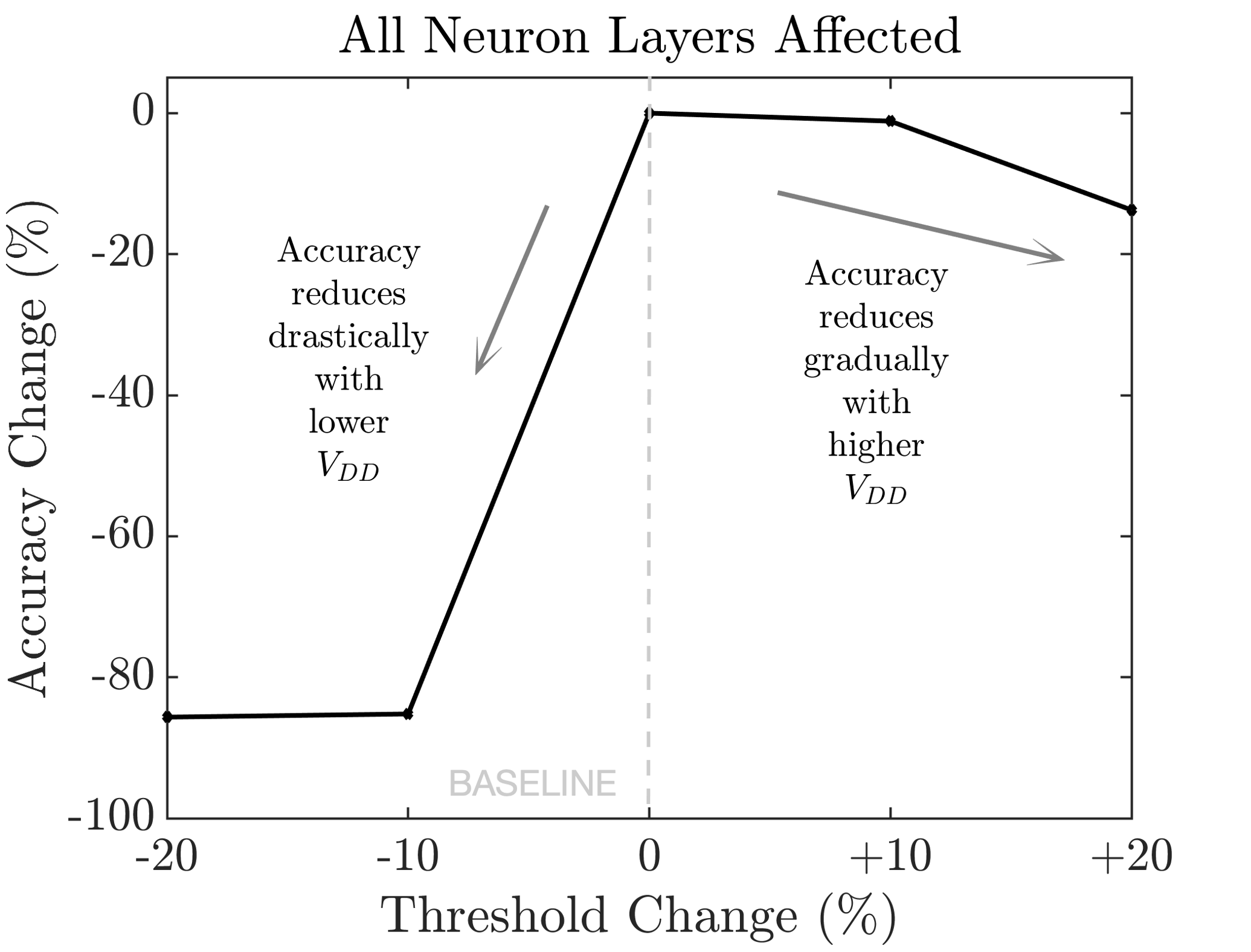}
                  \vspace{-11mm}
                \caption{}
                \label{EL_IH_both}
        \end{subfigure}%
        \vspace{-0mm}
        \caption{Classification accuracy trend with SNN membrane threshold for (a) Excitatory Layer only (Attack 2); (b) Inhibitory Layer only (Attack 3); and (c) Both Excitatory Layer and Inhibitory Layer (Attack 4);}
        \label{snac2_sims}
        \vspace{-3mm}
\end{figure*}

\begin{figure*} [t] 
        \centering 
        (a)
        \begin{subfigure}[b]{0.27\textwidth}
                \centering
                \includegraphics[width=0.99\linewidth]{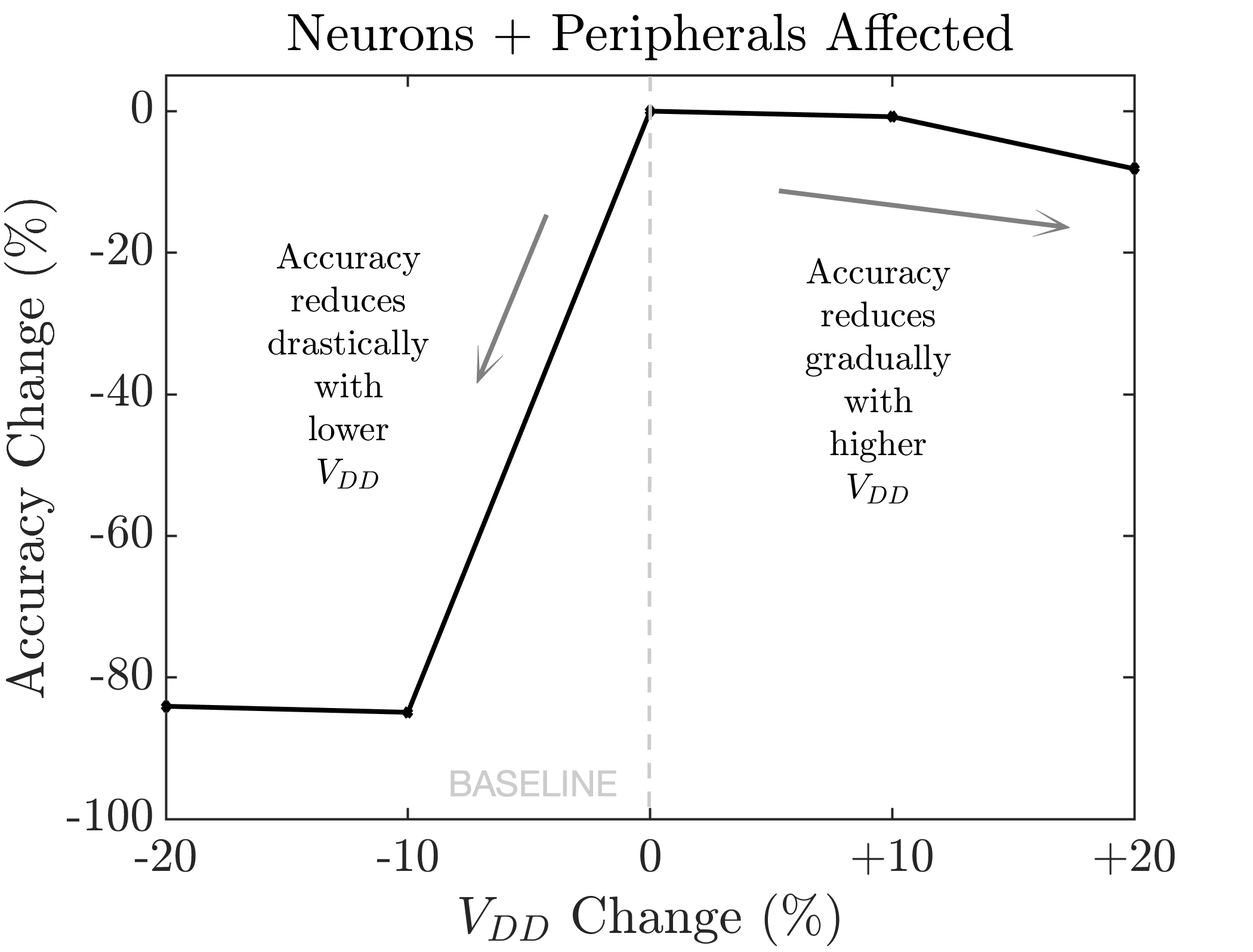}
                \vspace{-11mm}
                \caption{}
                \label{all_affected}
        \end{subfigure}%
        \hspace{0mm} (b)
        \hspace{0mm}
        \begin{subfigure}[b]{0.24\textwidth}
                \centering
                \includegraphics[width=0.99\linewidth]{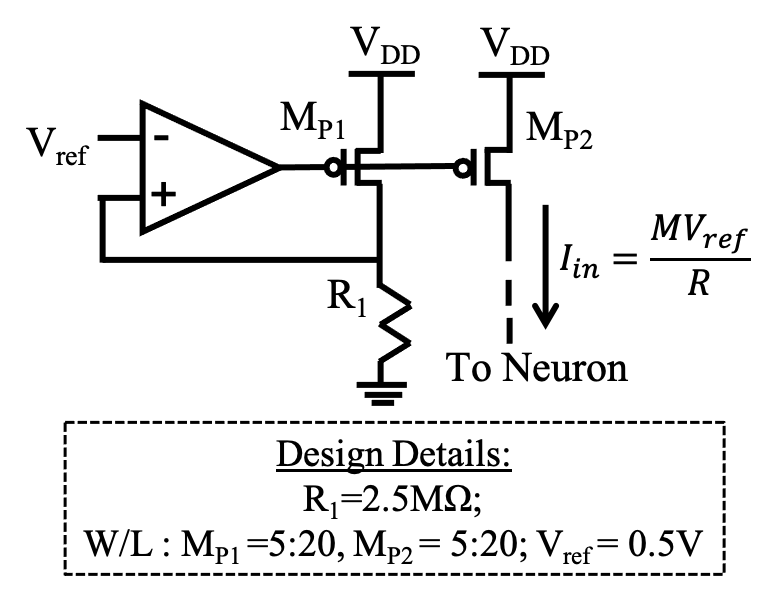}
                \vspace{-9mm}
                \caption{}
                \label{def_cur_driver}
        \end{subfigure}%
         \hspace{0mm} (c)
        \hspace{0mm} 
        \begin{subfigure}[b]{0.27\textwidth}
                \centering
               \includegraphics[width=0.99\linewidth]{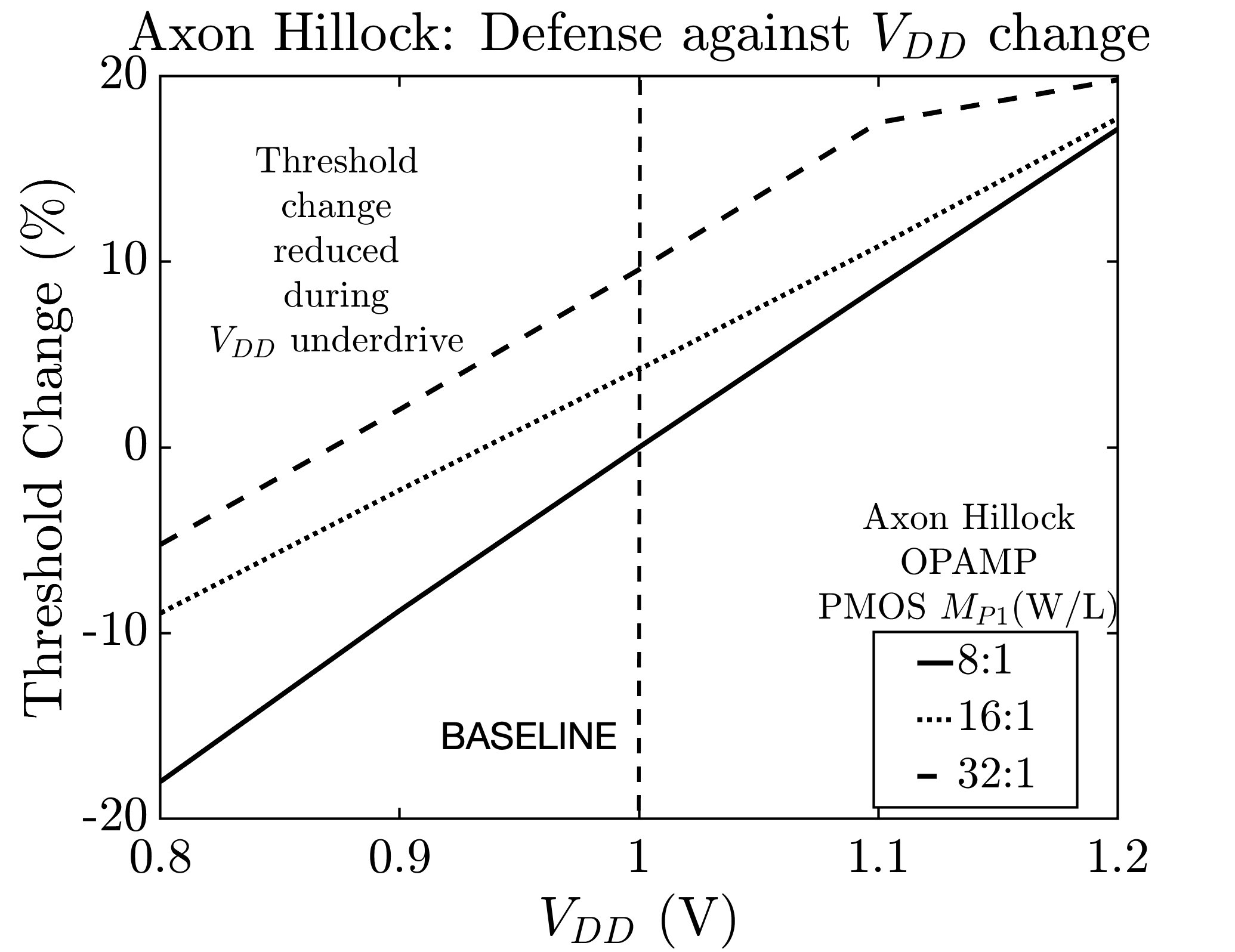}
                 \vspace{-11mm}
                \caption{}
                \label{axon_fixed}
        \end{subfigure}%
        \vspace{-0mm}
        \caption{(a) Change in classification accuracy with $V_{DD}$ change for entire system (Neurons and peripherals); (b) Robust SNN current driver (constant output spike amplitude); and (c) Axon Hillock Neuron: Effect of $M_{P1}$'s W/L on threshold voltage change during $V_{DD}$ manipulation. }
        \label{snapb_defense}
        \vspace{-3mm}
\end{figure*}

 \begin{figure*} [t] 
        \centering 
        (a)
        \begin{subfigure}[b]{0.18\textwidth}
                \centering
                \includegraphics[width=0.99\linewidth]{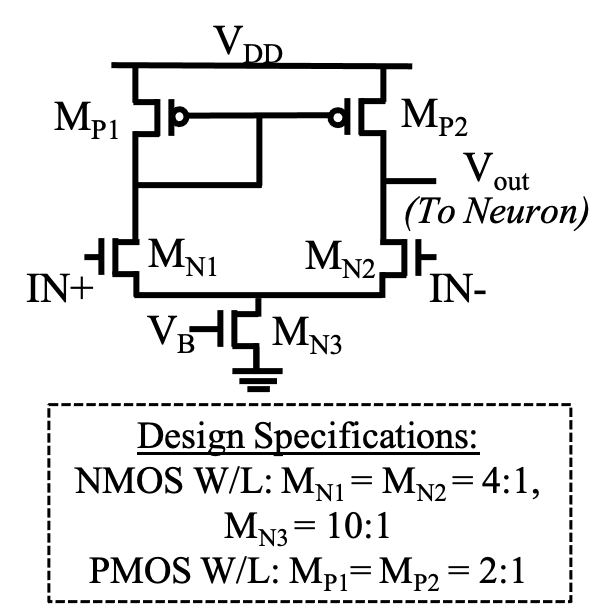}
                \vspace{-8mm}
                \caption{}
                \label{comparator}
        \end{subfigure}%
        \hspace{0mm} (b)
        \hspace{0mm}
        \begin{subfigure}[b]{0.36\textwidth}
                \centering
                \includegraphics[width=0.99\linewidth]{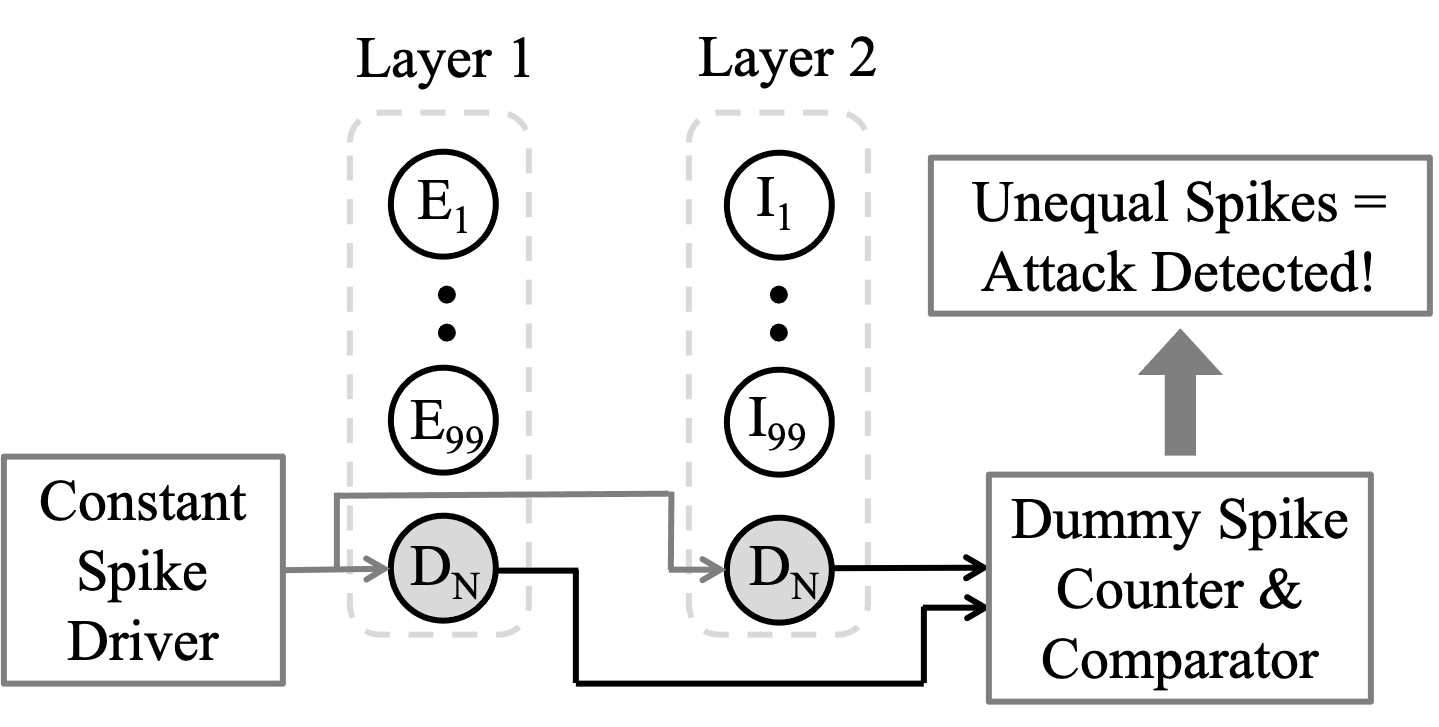}
                \vspace{-5mm}
                \caption{}
                \label{dummy_neuron}
        \end{subfigure}%
         \hspace{0mm} (c)
        \hspace{0mm} 
        \begin{subfigure}[b]{0.27\textwidth}
                \centering
               \includegraphics[width=0.99\linewidth]{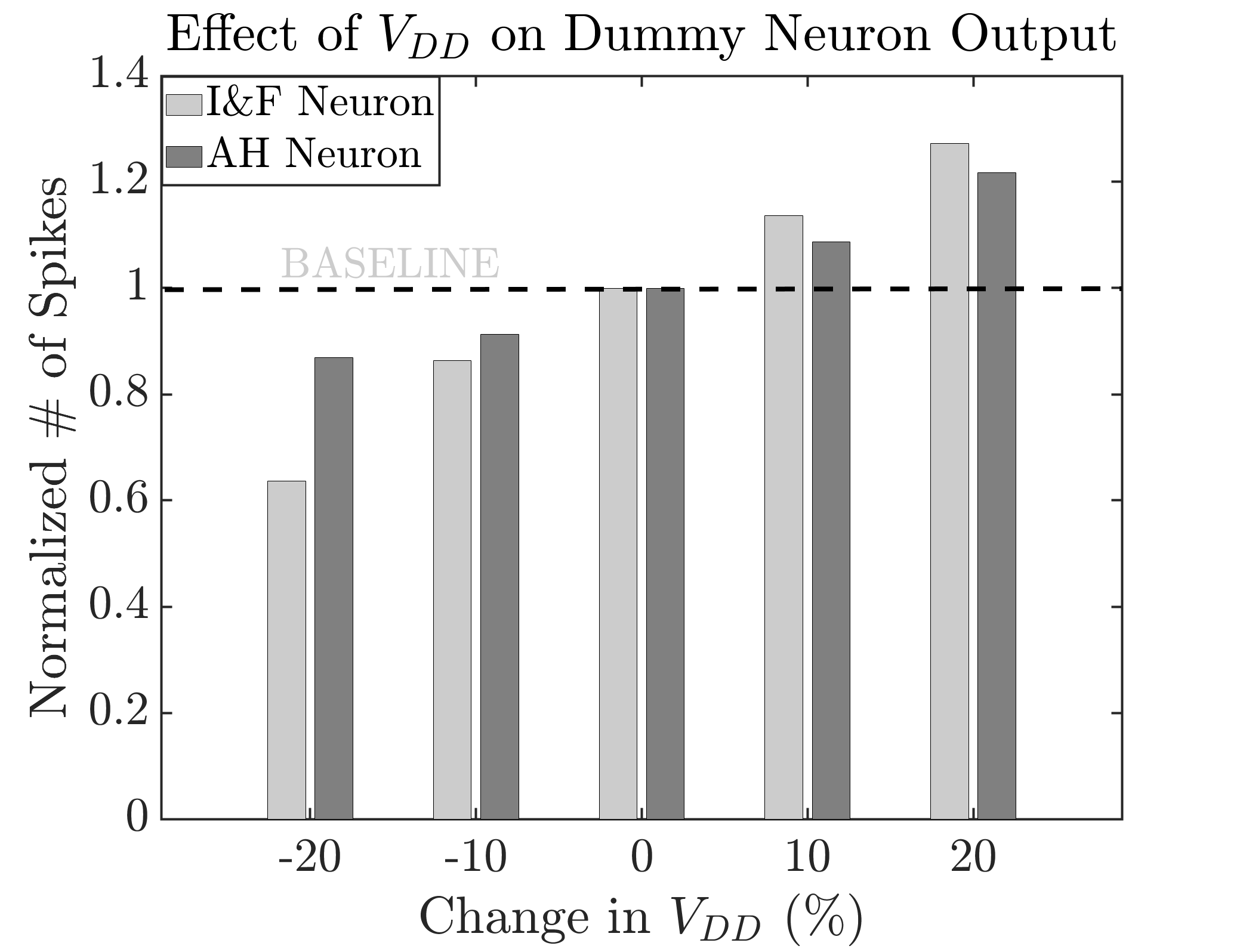}
                 \vspace{-11mm}
                \caption{}
                \label{dummy_neuron_sim}
        \end{subfigure}%
        \vspace{-0mm}
        \caption{(a) Comparator designed and implemented in the Axon Hillock Neuron to mitigate threshold variation.; (b) $V_{DD}$ change detection using dummy neuron; and (c) Effect of $V_{DD}$ on dummy neuron output. }
        \label{new_defense}
        \vspace{-5mm}
\end{figure*}

\subsection{SNN Threshold Manipulation  } \label{WB-2}
In Section \ref{snac2_des}, it is shown that the adversary can manipulate the membrane threshold voltages of the SNN neurons. This manipulation of the membrane threshold voltage ($V_{thr}$) affects classification accuracy. 
We have manipulated the threshold values of neurons in each layer from -20\% to +20\% to introduce power attacks in the BindsNET SNN. This range has been selected in line with the threshold variation observed in Fig. \ref{snn_thresh_vdd}. The change in threshold has different effects on neurons from the EL versus IL. Therefore, we analyze individual effect of each neuron layer's on classification accuracy. Finally, we analyze the response for all the layers on the classification accuracy. Note, Attacks 2 to 5 are \emph{white box} attacks since the adversary requires the location of the individual SNN layers (can be obtained from the layout) to induce the localized faults.

\textbf{Attack 2:} In this case, we subject only the EL to membrane threshold variation to study its individual effect on classification accuracy. This attack is possible when, (i) each neuron layer has their own voltage domain and the adversary injects a laser-induced fault, (ii) neuron layers share voltage domain but the local fault injection in one layer does not propagate to other layers due to the capacitance of the power rail. Various fraction of neurons in this layer, ranging from 0\% to 100\% are subject to -20\% to +20\% threshold change 
This analysis is performed to model the situation when adversary has fine grain control of the $V_{DD}$ inside a voltage domain e.g., using local voltage glitching attack that affects only a section of neurons. This is possible in systems which have thousands of neurons per layer that may be physically isolated due to interleaving synapse arrays. Fig. \ref{EL_only} shows the corresponding change in the classification accuracy. It is noted that classification accuracy is equal to or better than the baseline accuracy for threshold changes as long as $\leq$90\% of the layer is affected. For the worst case threshold change of -20\%, the classification accuracy degrades by 7.32\% when 100\% of the EL is affected. 
In summary, attacking the EL alone has a relatively low impact on the output accuracy. This is intuitive since the effect of any corruption in the EL can be recovered in the following IL. 

\textbf{Attack 3:} In this attack, we subject only the IL to membrane threshold change. Various fraction of neurons in this layer ranging from 0\% to 100\%, are subject to -20\% to +20\% threshold change. Fig. \ref{IH_only} shows the corresponding change in classification accuracy. It is noted that classification accuracy degrades below the baseline accuracy for 3 out of 4 cases of threshold change and for all fractions of IL affected. A worst case degradation of 84.52\% below the baseline accuracy (observed at -20\% threshold change at 100\% of IL affected) is noted. 
In summary, attacking the IL has a more significant effect on output accuracy compared to attacking the EL alone. This is understandable since IL is the final layer before the output. Therefore, any loss in learning cannot be recovered. 

\textbf{Attack 4:} In this attack, we subject 100\% of both the EL and the IL to the same membrane threshold change. Fig. \ref{EL_IH_both} shows the variation in accuracy with the threshold for both the layers of neurons. It is seen that the classification accuracy falls sharply as the membrane threshold of both the layers decreases below the baseline. A worst case accuracy degradation of 
-85.65\% below baseline accuracy is observed when the membrane threshold is reduced by 20\%.

\subsection{Input Spike Corruption and Threshold Manipulation} \label{BB}
\textbf{Attack 5: } This is a \emph{black box} attack where the adversary does not need to know the internal architecture of the current driver or the SNN neurons. Here we assume that the power supply is shared among all the components of the SNN system, including the current drivers and all of the neuron layers. Manipulating the $V_{DD}$ changes both membrane voltage per spike (\emph{theta}) and the threshold voltages ($V_{thr}$) of the SNN neurons. Fig. \ref{all_affected} shows that the worst case accuracy degradation is -84.93\%.

 \subsection{Summary of Power Attack Analysis}
From our analysis, we conclude following: 
 \subsubsection{SNN Assets} These include: (a) spike rate and amplitude, (b) neuron membrane threshold, (c) membrane voltage change per spike. Other assets (not studied in this paper) are strength of synaptic weights between neurons and the SNN learning rate. 
\subsubsection{SNN Vulnerabilities} $V_{DD}$ manipulation, (a) generation of spikes of lower/higher amplitude than nominal value by the neuron's input current driver, (b) lowers/increases neuron's membrane threshold. Both these vulnerabilities cause affected neurons to spike faster/slower. 

\subsubsection{Attack Models} Manipulation of global and local fine-grained power supply corrupts critical training parameters. 

Attacks not covered in this paper are, (a) generation of adversarial input samples to cause mis-classification, (b) fault injection into synaptic weights, (c) noise injection in input samples to attack specific neurons.

\section{Defenses}

\subsection{Robust Current Driver Design}
We propose a current driver that produces neuron input spikes of constant amplitude (Fig. {\ref{def_cur_driver}}). Here the negative input terminal of the op-amp is forced to a reference voltage that leads the positive terminal to be virtually connected to the reference voltage ($V_{Ref}$). The current through {$M_{P1}$} transistor is {$V_{Ref}/R_{1}$} and the negative feedback of the amplifier forces the gate voltage of {$M_{P1}$} to satisfy the current equation of the transistor. Since {$V_{GS}$} and {$V_{th}$} of {$M_{P1}$} and {$M_{P2}$} transistors are same, {$M_{P2}$} passes the same current as {$M_{P1}$}. 
Note, we have used long channel transistors to reduce the effect of channel length modulation. The power overhead incurred for the proposed robust current driver compared to the unsecured version is 3\%. Note that the area overhead of robust driver is negligible compared to the area of unsecured driver since the neuron capacitors occupy the majority of the area.



\subsection{Resiliency to Threshold Voltage Variation}

\subsubsection{Voltage Amplifier I\&F Neuron} 
In order to prevent $V_{thr}$ from being corrupted due to $V_{DD}$ change, it can be generated using a bandgap voltage reference that produces a constant voltage irrespective of power and temperature variations. A bandgap circuit is proposed in {\cite{sanborn2007sub}} that generates a constant $V_{ref}$ signal with a output variation of +/-0.56\% for supply voltages ranging from 0.85V to 1V at room temperature. A similar design can be used for our proposed I\&F neuron that require a constant external $V_{thr}$ signal. Since the $V_{thr}$ variation (+/-0.56\%) under $V_{DD}$ manipulation is negligible, the classification accuracy degradation reduces to $\sim$0\%. For our experimental 200-neuron implementation, the area overhead incurred by the bandgap circuit is 65\%. But this can be significantly reduced if the banggap circuit is shared with other components of the chip and if the SNNs are implemented with 10s of thousands of neurons as required by various applications.

\subsubsection{Axon Hillock Neuron} We propose following approaches,~\\
\textbf{Neuron transistor sizing: } In case of the Axon Hillock neuron (Fig. \ref{ind_neurons}a), the membrane threshold is determined by the $V_{DD}$ and the design of the first inverter (transistors $M_{P1}$ and $M_{N3}$). Simulations indicate that classification accuracy is affected mostly by lowering the membrane threshold as shown in Fig. \ref{EL_IH_both}. 
We increased the sizing of the PMOS transistor $M_{P1}$ to limit the threshold change due to $V_{DD}$. 
Fig. \ref{axon_fixed} shows that increasing the W/L ratio mitigates the reduction in threshold changes under lower $V_{DD}$. At 0.8V, the threshold change observed for W/L ratio of 32:1 is -5.23\% compared to -18.01 \% for the baseline sizing. The corresponding degradation in classification accuracy at $V_{DD}=0.8V$ is only 3.49\% which is a significant improvement compared to the 85.65\% degradation observed previously. At $V_{DD}=1.2V$, the threshold change increases by 3.2\% for W/L ratio of 32:1 and the corresponding accuracy degradation only increases by 1.4\%. For the upsized neuron, the power overhead observed is 25\% while the area overhead is negligible since majority of neuron area is occupied by the two 1pF capacitors that remain unchanged in the new design.

\textbf{Comparator implementation:} We replace the first inverter in the Axon Hillock neuron with a comparator that employs $V_{thr}$ generated by a bandgap circuit {\cite{sanborn2007sub}} as the reference voltage to eliminate the effect of $V_{DD}$ variation on inverter switching threshold. The rest of the design remains the same. Fig. {\ref{comparator}} shows the implemented comparator which ensures that the threshold voltage is not determined by the sizing of the inverter transistors or the $V_{DD}$. Instead, it depends on the input biasing of the proposed design. The IN+ and IN- bias is set to 600mV and $V_B$ is set to 400mV. The power overhead incurred is 11\% and the area overhead is negligible since the 1pF capacitors occupy majority of the neuron area.   




\subsection{Detection of $V_{DD}$ Change}
In addition to robust neuron design, we also propose a technique to voltage glitching attack directed at an individual neuron layer. This is done by introducing a dummy neuron within each neuron layer (shown in Fig. {\ref{dummy_neuron}}). In our design, the input of the dummy neuron is connected to a current driver that constantly drives spiking inputs of 200nA amplitude and spike width of 100ns. The spikes repeat every 200ns and does not depend on the spiking of the neurons from the previous layer. Under ideal conditions, the number of output spikes for a fixed sampling period for each dummy neuron should be identical. Fig. {\ref{dummy_neuron_sim}} shows the effect of of $V_{DD}$ change on the dummy neuron's output for both the I\&F and AH neurons over a sampling period of 100ms. It is seen that for both neurons, the number of dummy output spikes differs by $\geq 10\%$ as compared to the baseline. Note that this method is only effective against localized $V_{DD}$ change. For the SNN implemented in Section {\ref{SNN_model}}, the area and power overhead for the proposed dummy neuron detection mechanism is $\sim$ 1\% each.

\balance

\section{Conclusions}
We propose one \emph{black box} and {four} \emph{white box} attacks against commonly implemented SNN neuron circuits by manipulating its external power supply or inducing localized power glitches. We have demonstrated power-oriented corruption of critical SNN training parameters. 
We introduced the attacks for SNN-based digit classification tasks as test cases and observed significant degradation in classification accuracy. Finally, we proposed defenses against the proposed power-based attacks.

\section*{Funding}
This work is supported by SRC (2847.001 and 3011.001) and NSF (CNS-1722557, CCF-1718474, DGE-1723687, DGE-1821766, OIA-2040667 and DGE-2113839).

\bibliographystyle{ieeetr}
\bibliography{main}

\end{document}